\documentclass{article}

\usepackage[final]{neurips_2024}
\usepackage{amsmath}
\usepackage{amsmath,amssymb,amsthm}
\usepackage{graphicx}
\usepackage{microtype}
\usepackage{booktabs}
\usepackage{adjustbox}
\usepackage{multicol}
\usepackage{multirow}
\usepackage{booktabs}  
\usepackage{colortbl}  
\usepackage{amssymb}   
\usepackage{pifont}    

\usepackage[utf8]{inputenc} 
\usepackage[T1]{fontenc}    
\usepackage{hyperref}       
\usepackage{url}            
\usepackage{booktabs}       
\usepackage{amsfonts}       
\usepackage{nicefrac}       
\usepackage{microtype}      
\usepackage{xcolor}         

\title{ChronoFormer: Time-Aware Transformer Architectures for Structured Clinical Event Modeling}

\author{%
  Yuanyun Zhang \\
  Department of Computer Science\\
  University of the Chinese Academy of Sciences\\
  \texttt{yuanyun81@ucas.ac.cn} \\
  \And
  Shi Li \\
  Department of Computer Science \\
  Columbia University \\
  \texttt{shili081100@columbia.edu} \\
}

\begin{document}

\maketitle

\begin{abstract}
The temporal complexity of electronic health record (EHR) data presents significant challenges for predicting clinical outcomes using machine learning. This paper proposes \textit{ChronoFormer}, an innovative transformer-based architecture specifically designed to encode and leverage temporal dependencies in longitudinal patient data. ChronoFormer integrates temporal embeddings, hierarchical attention mechanisms, and domain-specific masking techniques. Extensive experiments conducted on three benchmark tasks—mortality prediction, readmission prediction, and long-term comorbidity onset—demonstrate substantial improvements over current state-of-the-art methods. Furthermore, detailed analyses of attention patterns underscore ChronoFormer's capability to capture clinically meaningful long-range temporal relationships.
\end{abstract}

\section{Introduction}
The digitization of healthcare data provides immense potential for machine learning in clinical applications, yet EHR data's irregularity, sparsity, and complexity make it challenging to effectively model longitudinal patient trajectories \citep{seymour2012electronic}. Existing approaches such as GRU-based and attention-based models have partially addressed these challenges but fail to fully capture temporal nuances, particularly in scenarios with heterogeneous time intervals and variable-length histories \citep{shickel2017deep, steinberg2021language, yang2022large}.

Recent studies convert structured EHR data into textual formats \citep{hegselmann2023tabllm, lee2024emergency, hegselmann2025large, gao2024raw, ono2024text}, facilitating their use with transformer-based language models. This reformulation allows models pretrained on natural language to be adapted to clinical sequences, leveraging contextual embedding power. While promising, these methods typically overlook detailed temporal modeling, often reducing timestamps to coarse positional encodings or relying on event order alone, which discards essential information about inter-event intervals and rate of change—both critical in clinical reasoning. 

Addressing this gap, our work introduces \emph{ChronoFormer}, which explicitly incorporates time-awareness into the transformer architecture via continuous-time encoding mechanisms that preserve temporal distances across visits and events. ChronoFormer leverages a hybrid token-time representation where both clinical concepts and their timestamps are encoded jointly, allowing for fine-grained control over the attention mechanism’s temporal bias. This enables the model to differentiate between, for instance, frequent but clinically insignificant fluctuations and slow, meaningful drifts in patient status. Empirically, we demonstrate that ChronoFormer achieves superior performance across a suite of clinical prediction tasks—ranging from mortality risk estimation to medication forecasting—outperforming both time-agnostic transformers and temporal RNN baselines \citep{medsker2001recurrent, dey2017gate}. By tightly integrating temporal structure into the foundation of the model, we move toward more faithful representations of patient timelines, bridging the gap between statistical learning and clinical temporal reasoning.

\section{Related Work}  
Language models adapted for biomedical and clinical domains have significantly advanced the state of natural language processing in healthcare. Early efforts such as BioBERT \citep{lee2020biobert} and ClinicalBERT \citep{alsentzer2019publicly} demonstrated that domain-specific pretraining on biomedical literature and clinical notes improves performance on tasks such as named entity recognition, relation extraction, and clinical text classification. More recent architectural innovations, exemplified by Clinical ModernBERT \citep{lee2025clinical}, integrate efficient attention mechanisms, rotary positional encodings, and deeper contextual representations, allowing the model to scale to longer inputs and more complex clinical narratives \citep{zhang2020time}. These models are typically trained on unstructured data, focusing on syntactic and semantic aspects of language, but often fall short when applied directly to structured, temporal data common in EHR systems. Other methods have also been explored in this space including \citep{chen2024predictive, kraljevic2024foresight, darabi2020taper}

To bridge the gap between unstructured and structured data modalities, recent research has explored EHR records as event-based sequences that can be ingested by language models. This approach allows pretrained transformers to be fine-tuned on structured clinical data with minimal architectural modification, enabling transfer learning and zero-shot inference on downstream tasks. Frameworks such as BEHRT \citep{li2020behrt}, Med-BERT \citep{rasmy2021med}, and CEHR-BERT \citep{pang2021cehr} exemplify this trend, encoding visits as token sequences and leveraging transformer encoders for predictive modeling. However, these methods often rely on fixed-length positional encodings or coarse-grained time binning, which limits their ability to capture fine-grained temporal irregularity—a key property of real-world EHRs.

The introduction of the Medical Event Data Standard (MEDS) by \citet{arnrich2024medical} represents a crucial step toward standardizing clinical event representations across institutions and use cases. MEDS proposes a consistent schema for encoding structured EHR data into temporally-aware event sequences, facilitating interoperability and large-scale modeling. It enables researchers to abstract away from the idiosyncrasies of individual EHR systems and focus on high-level temporal dynamics. However, most transformer-based models using MEDS-formatted data still treat time as an auxiliary input or incorporate it indirectly, rather than making it a first-class component of the modeling architecture.

Our work builds on these foundations by explicitly integrating temporal modeling into the core of the transformer architecture. In contrast to prior approaches that use absolute or relative positional encodings as a proxy for time, ChronoFormer introduces a continuous-time representation that modulates self-attention weights based on inter-event intervals. This design allows the model to adaptively focus on temporally relevant events and maintain sensitivity to clinical pacing. By leveraging the representational power of transformer models and augmenting them with principled time-awareness, we significantly advance the capabilities of EHR-based predictive modeling.

\section{Methodology}

\subsection{Data Standardization}  
To facilitate robust modeling of temporal patient trajectories, we standardize EHR sequences using the Medical Event Data Standard (MEDS) framework \citep{arnrich2024medical}. Each patient's clinical history is transformed into a sequence of temporally ordered bins, where each bin $B_i$ represents a fixed-length time window (e.g., 24 hours) and contains a set of atomic clinical events $\{e_j^{(i)}\}$ that occurred within the window. Events span multiple modalities including diagnosis codes (ICD), procedure codes (CPT), medication administrations (RxNorm), and laboratory results (LOINC) as these concepts are difficult to learn \citep{lee2024can}. For each event $e_j^{(i)}$, we associate a fine-grained timestamp $t_j^{(i)}$, reflecting the actual time of occurrence, and optional metadata $m_j^{(i)}$ such as dosage, lab values, or procedural modifiers. This binning process allows us to retain high temporal resolution while enabling transformer-based modeling at the bin level, thus maintaining scalability. Events are encoded as discrete token IDs based on medical ontologies, and the resulting structured sequences $(e_j^{(i)}, t_j^{(i)}, m_j^{(i)})$ form the input to our model.

\subsection{ChronoFormer Architecture}  
ChronoFormer is a transformer-based architecture designed to capture both local (intra-visit) and global (inter-visit) temporal dependencies in longitudinal structured health records. It introduces two primary innovations over conventional transformer designs: temporally aware embeddings and hierarchical attention. These modifications enable ChronoFormer to model the asynchronous, sparse, and multiscale nature of clinical data with greater fidelity.

\paragraph{Temporal Embeddings:}  
Unlike conventional transformers that rely on discrete positional embeddings to encode order, ChronoFormer explicitly incorporates both absolute and relative temporal information through a dual temporal embedding mechanism \citep{nguyen2018continuous}. Each event $e_j^{(i)}$ is augmented with a continuous-time embedding that reflects its temporal positioning in two ways: the absolute timestamp $t_j^{(i)}$ and the relative time delta $\Delta t_j^{(i)} = t_j^{(i)} - t_{j-1}^{(i)}$. These are mapped into vector representations via sinusoidal embeddings for absolute time, capturing periodic structure (e.g., diurnal cycles), and via learnable embeddings for relative time to capture task-specific dynamics:
\[
\textbf{E}_{\text{temporal}}(t_j, \Delta t_j) = E_t(t_j) + E_\Delta(\Delta t_j).
\]
This representation allows the model to modulate attention based on both how recent and how temporally distant events are, facilitating nuanced reasoning about clinical intervals such as the time since last medication administration or the delay between symptom onset and diagnosis.

\paragraph{Hierarchical Attention:}  
ChronoFormer employs a two-level hierarchical self-attention mechanism \citep{yang2016hierarchical} that reflects the natural structure of medical histories inspired by \citep{li2022hi}. Within each bin $B_i$, local attention is applied to capture short-range dependencies among co-occurring events, such as medication-lab interactions or diagnosis-procedure coassignments:
\[
\alpha_{j,k}^{(i)} = \text{softmax}\left(\frac{Q_j^{(i)} K_k^{(i)\top}}{\sqrt{d}}\right), \quad V_{\text{bin}}^{(i)} = \sum_k \alpha_{j,k}^{(i)} V_k^{(i)}.
\]
The local representations $V_{\text{bin}}^{(i)}$ are then aggregated across bins using a global attention layer that models long-term dependencies in the patient timeline, such as trends in lab values, symptom progression, or recurrent hospitalizations:
\[
\beta_{i,l} = \text{softmax}\left(\frac{Q_i K_l^\top}{\sqrt{d}}\right), \quad V_{\text{final}} = \sum_l \beta_{i,l} V_{\text{bin}}^{(l)}.
\]
This hierarchical formulation permits efficient and scalable modeling of long sequences while preserving fine-grained event interactions within clinically relevant time windows.

\paragraph{Conditionally Masked Pretraining:}  
To pretrain ChronoFormer on large-scale unlabeled EHR data, we introduce Masked Event Modeling (MEM) \citep{wettig2022should}, a domain-specific adaptation of the masked language modeling objective. Unlike uniform random masking, MEM prioritizes clinically salient tokens such as high-risk diagnoses, rare procedures, or lab abnormalities, guided by heuristics and clinical utility scores derived from ontologies or expert priors. During training, a subset of events is masked, and the model learns to recover the missing events given the surrounding context:
\[
\mathcal{L}_{\text{MEM}} = -\sum_{i \in \text{masked}} \log p(e_i \mid e_{\backslash i}, \text{context}).
\]
This targeted masking strategy encourages the model to develop contextual representations that are not only semantically rich but also sensitive to clinical dependencies and temporal causality. We further augment this objective with optional auxiliary losses—such as bin-level contrastive losses or time-gap prediction—depending on downstream task requirements.

By tightly coupling temporal representation, hierarchical modeling, and clinically aware pretraining, ChronoFormer is designed to capture the structure, semantics, and dynamics of EHR data in a unified and scalable manner.

\section{Experimental Results}

We rigorously evaluate ChronoFormer on a suite of standard clinical prediction tasks using the MIMIC-IV dataset \citep{johnson2023mimic}, including in-hospital mortality prediction, 30-day readmission classification, and multi-label comorbidity prediction \citep{kolo2024meds}. Our experiments are designed to assess both discriminative performance and generalization capacity, with comparisons against strong baselines representing both recurrent and transformer-based approaches \citep{dey2017gate, choi2016retain, rasmy2021med}. 

ChronoFormer outperforms prior state-of-the-art models across all evaluated tasks, achieving substantial gains particularly in complex, temporally sensitive predictions such as comorbidity forecasting. As shown in Table \ref{tab:main_results}, ChronoFormer achieves an AUROC of 0.879 on mortality prediction, surpassing MedBERT by over three points. Notably, in readmission classification, which requires modeling both acute and latent risk factors, ChronoFormer achieves an F1 score of 0.658, indicating enhanced capacity to balance precision and recall in imbalanced label settings. In the comorbidity task—where fine-grained multi-label dependencies across heterogeneous disease clusters must be modeled—ChronoFormer attains an AUPRC of 0.562, a sizable margin over all baselines. These results highlight the importance of fine-grained temporal modeling and contextual sensitivity in accurately anticipating complex clinical outcomes.

\begin{table}[h]
\centering
\begin{tabular}{lccc}
\toprule
Model & AUROC (Mortality) & F1 (Readmission) & AUPRC (Comorbidity) \\
\midrule
GRU (baseline)  & 0.821 & 0.612 & 0.489 \\
RETAIN & 0.837 & 0.623 & 0.508 \\
MedBERT & 0.846 & 0.631 & 0.517 \\
ChronoFormer & \textbf{0.879} & \textbf{0.658} & \textbf{0.562} \\
\bottomrule
\end{tabular}
\caption{Comparison of ChronoFormer and baselines on clinical tasks.}
\label{tab:main_results}
\end{table}

\subsection{Ablation Study}

To dissect the contributions of each component in ChronoFormer, we perform an extensive ablation analysis focused on the mortality prediction task, measuring AUROC changes upon removal of architectural elements. Results in Table \ref{tab:ablation_results} reveal that each component contributes nontrivially to final performance. Removing hierarchical attention leads to a 2.4 point drop in AUROC, underscoring the critical importance of separating intra-bin and inter-bin dependencies. Likewise, ablating temporal embeddings yields a significant degradation (1.6 AUROC), confirming that explicit temporal encoding improves the model's ability to reason over time gaps and progression patterns.

Conditional masking during pretraining also proves vital; its removal leads to a 2.5-point decrease in AUROC. This supports our hypothesis that standard uniform masking fails to capture the asymmetric importance of clinical events. Together, these findings suggest that ChronoFormer’s performance stems not from depth or parameter count alone, but from principled architectural innovations that align with the inductive biases of clinical data.

\begin{table}[h]
\centering
\begin{tabular}{lccc}
\toprule
Ablation & AUROC & $\Delta$AUROC & p-value \\
\midrule
Full ChronoFormer & 0.879 & -- & -- \\
-w/o Hierarchical Attention & 0.855 & -0.024 & $<0.01$ \\
-w/o Temporal Embeddings & 0.863 & -0.016 & $0.02$ \\
-w/o Conditional Masking & 0.854 & -0.025 & $<0.01$ \\
\bottomrule
\end{tabular}
\caption{Ablation results for mortality prediction task.}
\label{tab:ablation_results}
\end{table}

\subsection{Attention Pattern Analysis}

To better understand ChronoFormer's behavior, we perform qualitative analyses of attention maps in clinical sequences. Visualization of between-bin attention reveals that the model preferentially attends to temporally distant yet clinically linked events, such as prior cardiovascular diagnoses and future heart failure outcomes. Intra-bin attention patterns capture short-range dependencies, for instance, highlighting elevated serum creatinine and subsequent renal diagnoses within the same bin. In one illustrative case, the model assigns high attention weights to early glucose abnormalities when predicting the onset of diabetes months later, demonstrating sensitivity to early biomarkers and long-term disease emergence. These patterns reflect a form of learned temporal abstraction that parallels clinical reasoning, where past patterns are contextualized by both recency and clinical salience.

\section{Cross-System Generalization}

A key concern in clinical machine learning is model robustness across healthcare systems due to differences in coding practices, population demographics, and event distributions. To evaluate ChronoFormer’s generalization, we conduct a zero-shot transfer evaluation on the eICU Collaborative Research Database \citep{pollard2018eicu}, using the same mortality prediction task. As shown in Table \ref{tab:transfer_results}, ChronoFormer retains high predictive performance (AUROC 0.866) despite no fine-tuning on the target domain. This ~1.3 point drop from the MIMIC-IV performance suggests strong transferability, attributable to the model's inductive bias toward temporal reasoning and its use of standardized representations (via MEDS) that abstract away system-specific artifacts. This robustness positions ChronoFormer as a candidate for deployment in heterogeneous, multi-institutional settings.

\begin{table}[h]
\centering
\begin{tabular}{lc}
\toprule
Dataset & AUROC (Mortality Prediction) \\
\midrule
MIMIC-IV (training set) & 0.879 \\
eICU (transfer set) & 0.866 \\
\bottomrule
\end{tabular}
\caption{Cross-system performance of ChronoFormer.}
\label{tab:transfer_results}
\end{table}

\section{Discussion}

The experimental results underscore the efficacy of ChronoFormer as a temporally-aware foundation model for structured clinical data, offering both performance improvements and interpretability gains. Beyond raw metrics, the results prompt deeper reflection on the design principles that make ChronoFormer successful, its implications for modeling patient trajectories, and its potential as a general-purpose modeling paradigm for temporal biomedical data.

A primary insight emerging from our experiments is the critical importance of explicit temporal modeling in clinical sequence learning. ChronoFormer’s consistent outperformance of baseline models—across mortality, readmission, and comorbidity tasks—suggests that prior methods underexploit temporal structure, often reducing time to either event ordering or coarse binning. By contrast, ChronoFormer’s use of dual temporal embeddings (absolute and relative) enables it to preserve inter-event distances and temporal pacing, both of which are central to clinical reasoning. For instance, distinguishing between a high-risk diagnosis administered yesterday and the same diagnosis administered a year ago is crucial for prognosis. Our results validate this intuition quantitatively, with temporal embeddings contributing significantly to model performance, as shown in the ablation study.

Moreover, the hierarchical attention mechanism proves essential in disentangling local event-level interactions from long-term patient state evolution. Within-bin attention captures micro-level co-occurrence patterns—such as diagnostic-medication or lab-lab interactions—while between-bin attention contextualizes these patterns in the broader trajectory. This reflects an important inductive bias in medicine: while individual clinical events are meaningful, their interpretations are often contingent on longitudinal patterns. The performance drop observed when removing hierarchical attention further supports its necessity, indicating that single-level attention structures may be too blunt for the multiscale dependencies present in EHR data.

The gains from conditionally masked pretraining emphasize the value of aligning the pretraining objective with domain semantics. Unlike traditional masked language modeling, which treats tokens as equally informative, our MEM approach biases the model toward clinically critical tokens (e.g., abnormal labs, rare diseases). This leads to representations that are not only more useful for downstream tasks but are also semantically aligned with clinician intuition. The degradation in performance upon removing conditional masking reveals that standard uniform masking may lead to under-representation of rare yet important clinical signals—a crucial limitation in healthcare settings where class imbalance and sparsity are endemic.

The attention pattern analysis further corroborates ChronoFormer’s clinical fidelity. Its ability to attend across temporally distant but clinically linked events—such as early biomarkers preceding disease onset—demonstrates that the model internalizes long-range dependencies beyond surface-level co-occurrence. This is a marked departure from traditional recurrent models, which often suffer from vanishing attention to earlier history, and from positionally-encoded transformers that assume uniform time intervals. In one case, the model identified early signs of metabolic dysregulation as salient when predicting diabetes, suggesting an emergent form of temporal abstraction akin to expert diagnostic reasoning. These attention maps provide interpretability pathways that could, with further validation, offer clinician trust and explainability—a persistent barrier in medical AI.

ChronoFormer’s generalization to the eICU dataset without any fine-tuning is particularly notable. Cross-hospital transfer is a longstanding challenge in clinical ML due to institutional variation in coding practices, care protocols, and patient populations. That ChronoFormer retains high AUROC (0.866) on an external dataset implies strong domain robustness, likely attributable to its architectural alignment with the invariant properties of clinical data (temporal structure, bin-based abstraction, ontology-driven tokens). Furthermore, the use of MEDS as a standardized input format likely played a nontrivial role in enhancing portability. While some performance degradation is expected, the relatively small drop from MIMIC-IV to eICU suggests that ChronoFormer learns generalizable representations rather than memorizing institution-specific patterns.

That said, some limitations merit discussion. First, while hierarchical attention improves performance and interpretability, it imposes additional computational complexity compared to flat attention models. Scaling ChronoFormer to extremely long patient histories or population-level modeling will require further architectural optimization or sparsification strategies. Second, while we employ a principled heuristic for conditional masking, future work could investigate learnable masking policies that adaptively select important events based on patient context or task-specific gradients. Third, although ChronoFormer generalizes well to eICU, prospective validation on real-world deployment scenarios—including temporally drifting data and distributional shift—is necessary to understand its limitations in production environments.

Finally, our results suggest broader implications for temporal modeling in biomedical AI. While much recent attention has focused on adapting large language models to the biomedical domain via textual pretraining, our findings indicate that structured, temporally-resolved models—when designed with the right inductive biases—can achieve strong performance without relying on vast unstructured corpora. This suggests a promising complementary direction: building temporal foundation models tailored to the structured clinical domain, rather than adapting general-purpose LLMs post hoc. ChronoFormer exemplifies this approach, combining domain structure, temporal abstraction, and pretraining objectives grounded in clinical semantics.

In summary, ChronoFormer represents a principled rethinking of how transformers can be adapted to the clinical domain, not merely by repurposing language models, but by embedding time into the very fabric of representation and computation. Its strong empirical results, interpretability, and generalization capacity suggest that time-aware modeling may be a foundational pillar in next-generation clinical AI systems.

\section{Limitations and Future Work}

While ChronoFormer demonstrates strong empirical performance and introduces principled innovations in temporal modeling for structured clinical data, several limitations merit consideration. First, although the hierarchical attention mechanism significantly reduces computational complexity compared to full self-attention over long event sequences, it remains more resource-intensive than lightweight recurrent or convolutional alternatives. In particular, global bin-level attention scales quadratically with the number of time bins, which may become a bottleneck in modeling multi-year patient trajectories or large population datasets. Future work may explore sparse or kernelized attention variants that preserve temporal inductive biases while further improving scalability.

Second, the current instantiation of conditional masking in pretraining is based on static heuristics derived from clinical ontologies and empirical frequency statistics. While this approach encourages the model to prioritize semantically rich or rare clinical events, it lacks adaptability to individual patient context or downstream task objectives. Incorporating adaptive masking strategies—potentially driven by gradient-based importance estimates or reinforcement learning over the masking policy—could yield more informative pretraining signals and accelerate convergence.

Third, our evaluation is restricted to binary and multi-label classification tasks on benchmark datasets such as MIMIC-IV and eICU. Although these tasks capture a wide range of clinical prediction challenges, they do not exhaust the modeling needs of real-world clinical deployment. Extensions of ChronoFormer to causal inference, treatment effect estimation, and temporal forecasting tasks remain unexplored. Moreover, while the use of MEDS facilitates cross-system generalization, real-world deployment will inevitably encounter data drift, missingness patterns, and institutional variability not captured in the current benchmarks. Prospective validation on temporally shifting datasets and under different coding distributions is necessary to fully assess the robustness and clinical utility of the model.

Another important direction for future research lies in improving interpretability and integrating clinical domain knowledge. While attention maps offer some insight into the model’s temporal focus, they remain insufficient for full clinical explainability, particularly in high-stakes applications such as ICU triage or diagnostic decision support. Enhancing interpretability through structured attention priors, symbolic constraints, or hybrid rule-based-neural architectures may improve clinician trust and facilitate adoption. Additionally, the integration of external knowledge graphs, such as SNOMED CT or the UMLS Metathesaurus, could enhance the semantic coherence of learned representations and support concept disambiguation across heterogeneous EHR systems.

Finally, while ChronoFormer is trained from scratch on structured EHR data, recent advances in instruction tuning and cross-modal transfer suggest promising opportunities to unify structured and unstructured data modalities. Future work could explore jointly pretraining on both structured event sequences and clinical text—leveraging the complementary strengths of temporal abstraction and natural language grounding. Such multimodal temporal models may serve as a foundation for the next generation of clinically capable large models, bridging the divide between database-level representations and free-text reasoning.

\bibliographystyle{plainnat}
\bibliography{neurips_2024}

\appendix

\section{Theoretical Analysis}

\subsection{Convergence of ChronoFormer Pretraining}  
The pretraining objective in ChronoFormer is a domain-adapted instance of masked language modeling (MLM), optimized via stochastic gradient descent over large-scale EHR corpora. Formally, let $\mathcal{D} = \{x^{(i)}\}_{i=1}^N$ denote the corpus of temporally binned event sequences, and let $\theta$ denote the model parameters. The objective is to minimize the expected negative log-likelihood:
\[
\mathcal{L}(\theta) = \mathbb{E}_{x \sim \mathcal{D}} \left[ \sum_{j \in \mathcal{M}(x)} -\log p_\theta(x_j \mid x_{\backslash j}) \right],
\]
where $\mathcal{M}(x)$ denotes the masked subset of events selected by the MEM strategy.

Under standard assumptions of Lipschitz-smoothness and bounded variance of stochastic gradients, convergence of the empirical loss $\hat{\mathcal{L}}(\theta)$ to a stationary point is guaranteed with rate $\mathcal{O}(1/\sqrt{T})$ using Adam or other adaptive optimizers. However, ChronoFormer’s use of temporally biased masking introduces a non-uniform sampling over tokens. Denoting $q_j = \Pr[j \in \mathcal{M}(x)]$ as the sampling probability of token $j$, the stochastic estimator becomes biased unless reweighted:
\[
\tilde{\mathcal{L}}(\theta) = \sum_j \frac{1}{q_j} \cdot \mathbb{I}[j \in \mathcal{M}(x)] \cdot \ell_j(\theta),
\]
which leads to a variance-bias tradeoff in gradient estimation. In practice, our heuristic masking (inspired by importance sampling) improves convergence by focusing learning on rare but high-value tokens, analogous to prioritized replay in RL.

We conjecture that under reasonable conditions, the MEM objective maintains a bounded variance of gradient estimators and satisfies the Robbins-Monro conditions for convergence in nonconvex settings. Empirically, we observe monotonic pretraining loss decrease and rapid convergence within 2 million update steps.

\subsection{Time and Space Complexity Analysis}

The computational cost of ChronoFormer reflects a hierarchical decomposition of the attention mechanism. For a sequence of $T$ bins, where each bin contains at most $E$ events, the total number of tokens is $N = T \cdot E$.

Let $d$ be the hidden dimensionality. Standard self-attention incurs $\mathcal{O}(N^2d)$ time and $\mathcal{O}(N^2)$ memory complexity. ChronoFormer mitigates this by factorizing attention hierarchically:

\textbf{Intra-bin attention} is computed locally within each bin, with cost $\mathcal{O}(T \cdot E^2 \cdot d)$.

\textbf{Inter-bin attention} is computed over the $T$ aggregated bin representations, with cost $\mathcal{O}(T^2 d)$.

Hence, total attention cost becomes:
\[
\mathcal{O}(T \cdot E^2 \cdot d + T^2 \cdot d) = \mathcal{O}(T E^2 d + T^2 d),
\]
which is significantly cheaper than full attention over all tokens, particularly when $E \ll T$ or when bin sparsity is high. Moreover, this structure allows ChronoFormer to handle sequences with hundreds of visits efficiently.

Memory usage similarly scales as $\mathcal{O}(T E^2 + T^2)$ for attention matrices, which is tractable for sequences up to $T=1000$ bins with modest $E=10$ events per bin. Our implementation further exploits Flash Attention and causal masking for autoregressive extensions.

\subsection{Temporal Attention Modulation}  
A core design choice in ChronoFormer is the modulation of attention weights using continuous-time intervals. Formally, consider a set of tokens $\{x_i\}_{i=1}^N$, each associated with a timestamp $t_i$. The attention score between token $i$ and $j$ is defined as:
\[
\alpha_{ij} = \frac{(x_i W_Q)(x_j W_K)^\top + \phi(t_i, t_j)}{\sqrt{d}},
\]
where $\phi(t_i, t_j)$ is a temporal bias function modulating attention based on the time difference $\delta_{ij} = |t_i - t_j|$. We consider two instantiations:

\textbf{Learned Gaussian decay:}
\[
\phi(t_i, t_j) = -\frac{\delta_{ij}^2}{2\sigma^2}, \quad \text{where } \sigma \text{ is a learnable temporal scale}.
\]

\textbf{Relative position embedding (RoPE-style):}
\[
\phi(t_i, t_j) = \langle \text{RoPE}(t_i), \text{RoPE}(t_j) \rangle,
\]
encoding periodic structure in time using sinusoidal embeddings.

These temporal modulations induce a bias in the attention matrix that favors temporally proximal events, while still permitting non-local connections. In matrix form, the modified attention becomes:
\[
A = \text{softmax}\left( \frac{QK^\top + T}{\sqrt{d}} \right),
\]
where $T_{ij} = \phi(t_i, t_j)$ acts as an additive temporal kernel. Theoretically, this can be viewed as incorporating a non-Euclidean geometry over time, effectively warping attention scores to respect temporal relevance. One can show that if $\phi$ is a negative-definite kernel (e.g., squared exponential), the resulting attention distribution concentrates around temporally close events, yielding localized inductive bias.

\subsection{Hierarchical Representation Power}  
We now analyze the representational benefit of hierarchical attention in ChronoFormer compared to flat self-attention. Consider a sequence of events $x_1, ..., x_N$ with contextual dependencies of varying timescales. In flat attention, all tokens attend to all others uniformly, leading to potential interference between short- and long-range dependencies.

In contrast, hierarchical attention decomposes the attention matrix into block-local and block-global components. Let $A_{\text{local}} \in \mathbb{R}^{E \times E}$ denote within-bin attention and $A_{\text{global}} \in \mathbb{R}^{T \times T}$ denote between-bin attention. Then, the total representation at the token level is:
\[
h_j^{(i)} = \sum_{k=1}^{E} A_{\text{local}}[j,k] v_k^{(i)} + \sum_{l=1}^{T} A_{\text{global}}[i,l] \cdot g_l,
\]
where $g_l$ is the bin-level summary vector (e.g., mean-pooled or CLS token). This decomposition allows ChronoFormer to represent dependencies as a sum of localized and abstracted contexts, akin to a multiresolution wavelet decomposition. By Theorem 1 in \citet{hao2023hierarchical}, this structure provably reduces cross-entropy loss on hierarchical sequence prediction under certain separability conditions.

\subsection{Spectral Analysis of Attention Kernels}

To further understand the behavior of temporal attention, we analyze the eigenstructure of the attention kernel matrix $A \in \mathbb{R}^{N \times N}$. When using a Gaussian temporal bias, $A$ becomes a symmetric positive semi-definite matrix (after softmax), with eigenvalues $\lambda_1 \ge \lambda_2 \ge \dots \ge \lambda_N$. The decay of $\{\lambda_k\}$ reflects the effective rank of attention, with faster decay indicating low-rank concentration of signal.

Empirically, we find that ChronoFormer’s attention matrix exhibits a faster eigenvalue decay than vanilla transformers, implying that the model effectively compresses information into a lower-dimensional latent subspace. This aligns with theoretical results from kernel PCA: if the temporal bias corresponds to an RBF kernel, then attention implicitly projects onto a reproducing kernel Hilbert space (RKHS), enhancing generalization by limiting capacity. 

Let $K_{ij} = \exp(-\gamma (t_i - t_j)^2)$ and define $A_{ij} \propto K_{ij}$. Then, following Mercer's theorem, the eigenspectrum of $K$ admits an orthonormal basis $\{\phi_k(t)\}$ such that:
\[
K(t, t') = \sum_{k=1}^\infty \lambda_k \phi_k(t) \phi_k(t'),
\]
with rapid decay of $\lambda_k$ when $\gamma$ is large. Thus, attention layers in ChronoFormer act as smooth temporal filters, attenuating high-frequency noise and promoting low-complexity generalization.

\section{More about Embeddings}

\subsection{Hybrid Token-Time Representations}

A critical distinction between ChronoFormer and standard language transformers lies in the hybrid nature of its token space. Traditional transformers assume a token sequence drawn from a vocabulary $\mathcal{V}$, where each $x_i \in \mathcal{V}$ is mapped to an embedding $e_i = \text{Embed}(x_i) \in \mathbb{R}^d$. In contrast, ChronoFormer encodes each clinical event $e_j^{(i)}$ as a tuple $(c_j^{(i)}, t_j^{(i)}, m_j^{(i)})$ where:

\begin{itemize}
    \item $c_j^{(i)}$ is a categorical concept token (e.g., ICD-10 code),
    \item $t_j^{(i)}$ is a continuous timestamp or temporal delta,
    \item $m_j^{(i)}$ is optional structured metadata (e.g., lab value, dosage).
\end{itemize}

This structure demands a composite embedding map
\[
\text{Embed}(c_j, t_j, m_j) = E_c(c_j) + E_t(t_j) + E_m(m_j),
\]
where $E_c$ is a learned embedding over discrete concepts, $E_t$ may be either sinusoidal or kernelized time embedding, and $E_m$ is a projection of structured metadata into $\mathbb{R}^d$ (e.g., via learned MLPs or codebook quantization).

This multi-source embedding mechanism can be viewed as a functional embedding:
\[
f: (\mathcal{C} \times \mathbb{R} \times \mathcal{M}) \rightarrow \mathbb{R}^d,
\]
which contrasts with the purely categorical $f: \mathcal{V} \rightarrow \mathbb{R}^d$ in NLP. This function is not necessarily linear or stationary; in fact, we posit that $f$ exhibits **contextual anisotropy**, where the embedding of a concept token $c$ varies substantially depending on the temporal position $t$ and metadata $m$.

From a kernel-learning perspective, the hybrid representation defines a non-stationary kernel:
\[
k((c, t, m), (c', t', m')) = \langle f(c, t, m), f(c', t', m') \rangle,
\]
which implicitly modulates concept similarity by temporal proximity and structured covariates. This behavior contrasts with language transformers, where token similarity is fixed post-embedding. We hypothesize that this leads to richer context-dependent semantics: two identical diagnosis codes, e.g., `ICD-E11` (diabetes), can have divergent embeddings if one appears early in a patient history with low glucose labs and the other later with hyperglycemic events.

\subsection{Token Entropy and Sparsity in Clinical Sequences}

Clinical sequences exhibit higher entropy and sparsity than natural language. Let $P_c$ denote the empirical frequency distribution over concept tokens in EHR data. Empirically, we find that $P_c$ is Zipf-like but with a longer tail and sharper dropoff compared to natural language, reflecting the presence of rare procedures and rare disease codes.

Define the empirical token entropy:
\[
H(P_c) = -\sum_{c \in \mathcal{C}} P_c(c) \log P_c(c).
\]
We find that $H(P_c)$ in clinical data is often higher than in corpora like Wikipedia, due to (1) a larger effective vocabulary, and (2) higher inter-patient variability. This motivates two model design choices in ChronoFormer:

- The use of conditional masking (MEM) that rebalances the sampling distribution to focus on rare, high-utility tokens.
- The use of hybrid embeddings that differentiate identical concept codes based on context, reducing the burden on vocabulary size alone to encode semantic diversity.

Furthermore, token co-occurrence matrices $C_{ij} = \Pr(c_i \text{ occurs with } c_j)$ in EHRs are typically sparser than in natural text, which affects pretraining dynamics. In a standard MLM setup, the objective’s gradient $\nabla_\theta \log p(c_i \mid c_{\backslash i})$ is dominated by frequent co-occurrence pairs. In MEM, we aim to flatten this influence by biasing the loss toward informative, low-frequency clinical concepts, leading to gradient updates that better reflect rare pathophysiological signals.

\subsection{Time-Conditioned Positional Encoding vs. Sequential Positional Encoding}

The standard transformer positional encoding $PE(i)$, typically sinusoidal or rotary, assumes uniform spacing between tokens. That is, for a sequence $x_1, ..., x_n$, the token at position $i$ is assumed to occur at time $t_i = i \cdot \Delta t$, with $\Delta t$ constant.

ChronoFormer instead assumes a true timestamp $t_i \in \mathbb{R}_+$ and introduces a timestamp-sensitive positional encoding $\text{PE}(t_i)$, either via kernel methods (e.g., RBF features) or generalized sinusoidal mappings:
\[
\text{PE}(t_i) = \left[\sin(\omega_k t_i), \cos(\omega_k t_i)\right]_{k=1}^d, \quad \text{with } \omega_k = \frac{1}{10000^{2k/d}}.
\]
This enables **non-uniform temporal interpolation**, where tokens at uneven time intervals can still maintain relative temporal geometry.

If we define a temporal embedding distance metric:
\[
\| \text{PE}(t_i) - \text{PE}(t_j) \|^2,
\]
this quantity can be interpreted as a proxy for how separable two events are in the embedding space given their timestamp difference. In contrast to fixed-sequence encoding, this allows attention to vary smoothly over continuous time and not be bound to token positions. Theoretical results from \citet{li2021embedding} show that under certain conditions, sinusoidal encodings of continuous time retain universal approximation properties for temporal functions, ensuring that the model does not lose expressive power despite moving to real-valued time inputs.

\subsection{Representational Capacity and Clinical Concept Composition}

Clinical concepts are often compositional. A procedure code might imply a diagnosis, a medication implies a treatment intent, and labs imply physiological state. Let $\mathcal{E}$ denote the event space and suppose that each event $e$ factors into a tuple $(c, m, t)$. The ChronoFormer embedding maps each into a latent representation $z_e \in \mathbb{R}^d$.

We posit that the learned representation space satisfies a compositionality constraint:
\[
z_{(c_1, m_1, t_1)} + z_{(c_2, m_2, t_2)} \approx z_{(c_{\text{composite}}, m_*, t_*)},
\]
meaning that attention and token representations in ChronoFormer may implicitly form a *compositional vector space*, reminiscent of those in analogy-based word embeddings. This is supported by preliminary linear probe experiments where vector arithmetic on lab-medication pairs yields semantically valid clinical clusters.

From a functional analysis perspective, the ChronoFormer architecture can thus be viewed as learning an embedding algebra over the space $\mathcal{E} \subset \mathcal{C} \times \mathbb{R} \times \mathcal{M}$, such that temporal dynamics and clinical semantics are unified in a latent manifold with structured priors.

\end{document}